\documentclass[twocolumn,english,conference]{IEEEtran}
\usepackage[T1]{fontenc}
\usepackage[latin9]{inputenc}
\usepackage{graphicx}
\usepackage{subscript}

\makeatletter
\usepackage{url}

\@ifundefined{showcaptionsetup}{}{%
 \PassOptionsToPackage{caption=false}{subfig}}
\usepackage{subfig}
\makeatother

\usepackage{babel}
\begin{document}
\author{ 
\IEEEauthorblockN{John Papadakis, Andrew R. Willis} 
\IEEEauthorblockA{
	Department of Electrical and Computer Engineering\\
	University of North Carolina at Charlotte\\ 
	Charlotte, NC 28223-0001\\
	Email: arwillis@uncc.edu
}}
\IEEEoverridecommandlockouts 
\makeatletter 
\let\ORGps@IEEEtitlepagestyle\ps@IEEEtitlepagestyle \def\ps@IEEEtitlepagestyle{\ORGps@IEEEtitlepagestyle \def\@oddfoot{\hbox{}\@IEEEfooterstyle\footnotesize \raisebox{\footskip}[0pt][0pt]{\@IEEEpubid}\hss\hbox{}}\relax 
\def\@evenfoot{\@oddfoot}} 
\makeatother 
\IEEEpubid{978-1-5386-1539-3/17/\$31.00 ~\copyright2017 IEEE} 
\IEEEpubidadjcol

\title{Real-Time Surface Fitting to RGBD Sensor Data}
\maketitle
\begin{abstract}
This article describes novel approaches to quickly estimate planar
surfaces from RGBD sensor data. The approach manipulates the standard
algebraic fitting equations into a form that allows many of the needed
regression variables to be computed directly from the camera calibration
information. As such, much of the computational burden required by
a standard algebraic surface fit can be pre-computed. This provides
a significant time and resource savings, especially when many surface
fits are being performed which is often the case when RGBD point-cloud
data is being analyzed for normal estimation, curvature estimation,
polygonization or 3D segmentation applications. Using an integral
image implementation, the proposed approaches show a significant increase
in performance compared to the standard algebraic fitting approaches.
\end{abstract}

Real-time perception of scene geometry is an important problem to
the machine vision and robotics community. Solutions to this problem
play a crucial role in software systems that seek to endow machines
with a structural understanding of their environment. Applications
of these algorithms include scene navigation, interaction, and comprehension. 

Recently developed RGBD sensors have become popular for measuring
both scene appearance and geometry. These sensors provide high resolution
visual data in a low cost (\textasciitilde{}\$200USD) and compact
package. RGBD sensors combine a traditional color camera (RGB) with
an infrared depth sensor (D), and merge this data to produce HD color+range
images at real-time frame rates (\textasciitilde{}30 fps). Unlike
traditional 2D cameras, RGBD sensors provide depth measurements that
directly impart a sense of scene geometry, without the use of techniques
such as stereoscopic reconstruction. These sensors have seen increased
use as research tools for many computer vision related problems \cite{6547194}.

The volume of data produced by such sensors is quite large. As such,
real-time systems must carefully consider the computational cost of
algorithms that process this data. This is particularly important
for time sensitive tasks such as visual odometry-based navigation,
which relies on incoming sensor data to navigate geometrically complex
scenes.

For those RGBD algorithms that analyze the depth image, it is common
to extract geometric surfaces, e.g., planes and quadric surfaces,
from the sensed depth images. Fit surfaces often provide important
clues for both odometry, e.g., how the robot has moved, and for navigation,
i.e., identifying navigable regions of the scene. This article focuses
on planar surface fitting to point cloud data, a common task for geometric
scene analysis and segmentation algorithms such as surface normal
and curvature estimation, polygonization, and 3D segmentation \cite{Holz11real-timeplane}. 

This work describes a computationally efficient method for fitting
planar surfaces to RGBD data, suitable for real-time operation. This
formulation improves upon existing approaches by separating the traditional
least squares fitting problem into components including the intrinsic
camera parameters- which can be precomputed, and quantities related
to the disparity of the incoming depth data. When implemented using
integral images, this approach results in increased performance when
compared to the traditional formulation of these plane fitting methods.

\section{Prior Work \& Background Information}

It is necessary to discuss several results from prior work to provide
context for the contribution of this article and to lay preliminary
groundwork for the technical approach of the proposed algorithm. This
section includes a discussion of how $(X,Y,Z)$ point clouds are computed
from RGBD sensor data, the nature of RGBD sensor data noise, and a
discussion of leading approaches for 3D planar surface estimation
from point cloud data.

\subsection{Point Cloud Reconstruction}

Measured 3D $(X,Y,Z)$ positions of sensed surfaces can be directly
computed from the intrinsic RGBD camera parameters and the measured
depth image values. The $Z$ coordinate is directly taken as the depth
value and the $(X,Y)$ coordinates are computed using the pinhole
camera model. In a typical pinhole camera model, 3D $(X,Y,Z)$ points
are projected to $(x,y)$ image locations, e.g., for the image columns
the $x$ image coordinate is $x=f_{x}\frac{X}{Z}+c_{x}-\delta_{x}$.
However, for a depth image, this equation is re-organized to ``back-project''
the depth into the 3D scene and recover the 3D $(X,Y)$ coordinates
as shown by equation (\ref{eq:RGBD_reconstruction}) 

\begin{equation}
\begin{array}{ccc}
X & = & (x+\delta_{x}-c_{x})Z/f_{x}\\
Y & = & (y+\delta_{y}-c_{y})Z/f_{y}\\
Z & = & Z
\end{array}\label{eq:RGBD_reconstruction}
\end{equation}
where $Z$ denotes the sensed depth at image position $(x,y)$, $(f_{x},f_{y})$
denotes the camera focal length (in pixels), $(c_{x},c_{x})$ denotes
the pixel coordinate of the image center, i.e., the principal point,
and $(\delta_{x},\delta_{y})$ denote adjustments of the projected
pixel coordinate to correct for camera lens distortion.

\subsection{Measurement Noise}

Studies of accuracy for the Microsoft Kinect sensor show that a Gaussian
noise model provides a good fit to observed measurement errors on
planar targets where the distribution parameters are mean $0$ and
standard deviation $\sigma_{Z}=\frac{m}{2f_{x}b}Z^{2}$ for depth
measurements where $\frac{m}{f_{x}b}=-2.85e^{-3}$ is the linearized
slope for the normalized disparity empirically found in \cite{s120201437}.
Since 3D the coordinates for $(X,Y)$ are a function of both the pixel
location and the depth, their distributions are also known as shown
below:

\begin{equation}
\begin{array}{ccc}
\sigma_{X} & = & \frac{x-c_{x}+\delta_{x}}{f_{x}}\sigma_{Z}=\frac{x-c_{x}+\delta_{x}}{f_{x}}(1.425e^{-3})Z^{2}\\
\sigma_{Y} & = & \frac{y-c_{y}+\delta_{y}}{f_{y}}\sigma_{Z}=\frac{y-c_{y}+\delta_{y}}{f_{y}}(1.425e^{-3})Z^{2}\\
\sigma_{Z} & = & \frac{m}{f_{x}b}Z^{2}\sigma_{d'}=(1.425e^{-3})Z^{2}
\end{array}\label{eq:noise_models}
\end{equation}

These equations indicate that 3D coordinate measurement uncertainty
increases as a quadratic function of the depth for all three coordinate
values. However, the quadratic coefficient for the $(X,Y)$ coordinate
standard deviation is at most half that in the depth direction, i.e.,
$(\sigma_{X},\sigma_{Y})\approx0.5\sigma_{Z}$ at the image periphery
where $\frac{x-c_{x}}{f}\approx0.5$, and this value is significantly
smaller for pixels close to the optical axis.

\subsection{Implicit Plane Fitting to Point Cloud Data}

This section discussed the typical approach for 3D plane fitting which
estimated the plane when expressed as an implicit polynomial. We refer
to this approach as the implicit fitting method which seeks to minimize
the square of the perpendicular distance between $N$ measured $(X,Y,Z)$
data points and the estimated planar model, i.e.,

\begin{equation}
\epsilon(a,b,c,d)=\min_{a,b,c,d}\sum_{i=1}^{N}\left\Vert aX_{i}+bY_{i}+cZ_{i}+d\right\Vert ^{2}\label{eq:implicit_plane_fitting_error}
\end{equation}

We re-write this objective function as a quadratic matrix-vector product
by defining the vector $\mathbf{{\alpha}}=[\begin{array}{cccc}
a & b & c & d\end{array}]^{t}$ as the vector of planar coefficients and the matrix $\mathbf{{M}}$
as the matrix of planar monomials formed from the measured $(X,Y,Z)$
surface data having $i^{th}$ row $\mathbf{{M}}_{i}=[\begin{array}{cccc}
X_{i} & Y_{i} & Z_{i} & 1\end{array}]$. Using this notation, the optimization function becomes:
\[
\epsilon(\mathbf{{\alpha}})=\min_{\mathbf{\alpha}}\mathbf{{\alpha}}^{t}\mathbf{M}^{t}\mathbf{M}\mathbf{{\alpha}}
\]

Solving for the unknown plane equation coefficients requires a constraint
on the coefficient vector: $\left\Vert \mathbf{{\alpha}}\right\Vert ^{2}=\mathbf{{\alpha}}^{t}\mathbf{{\alpha}}=1$
to avoid the trivial solution $\mathbf{{\alpha}}=\mathbf{{0}}$. Equation
(\ref{eq:quadratic_error_w_lagrange_multiplier}) incorporates this
constraint as a Lagrange multiplier.
\begin{equation}
\epsilon(\mathbf{{\alpha}})=\min_{\mathbf{\alpha}}\left(\mathbf{{\alpha}}^{t}\mathbf{M}^{t}\mathbf{M}\mathbf{{\alpha}}-\lambda\left(\mathbf{{\alpha}}^{t}\mathbf{I}\mathbf{{\alpha}}-1\right)\right)\label{eq:quadratic_error_w_lagrange_multiplier}
\end{equation}

Taking the derivative of the error function then provides

\begin{equation}
\frac{d\epsilon(\mathbf{{\alpha}})}{d\mathbf{{\alpha}}}=\min_{\mathbf{\alpha}}\left(\left(\mathbf{M}^{t}\mathbf{M}-\lambda\mathbf{I}\right)\mathbf{{\alpha}}\right)\label{eq:derivative of error}
\end{equation}

Then from \cite{Fitzgibbon:1999:DLS:302943.302950,Taubin:Fitting:1991,Trucco:1998:ITC:551277}
the minimizer is known to be $\mathbf{\widehat{{\alpha}}}$, the eigenvector
associated with the smallest eigenvalue of the matrix $\mathbf{M}^{t}\mathbf{M}$
(also known as the scatter matrix). In general, $\mathbf{M}^{t}\mathbf{M}$
is a symmetric matrix and, for the monomials $\mathbf{{M}}_{i}=[\begin{array}{cccc}
X_{i} & Y_{i} & Z_{i} & 1\end{array}]$, the elements of this matrix are

\begin{equation}
\mathbf{M}^{t}\mathbf{M}=\sum_{i=1}^{N}\left[\begin{array}{cccc}
X_{i}^{2} & X_{i}Y_{i} & X_{i}Z_{i} & X_{i}\\
X_{i}Y_{i} & Y_{i}^{2} & Y_{i}Z_{i} & Y_{i}\\
X_{i}Z_{i} & Y_{i}Z_{i} & Z_{i}^{2} & Z_{i}\\
X_{i} & Y_{i} & Z_{i} & 1
\end{array}\right]\label{eq:XYZ_scattermatrix}
\end{equation}

Fitting implicit planar surfaces to point cloud data has become the
de-facto standard for point cloud processing algorithms and is now
part of many standard point cloud and image processing libraries,
e.g., OpenCV and PCL (Point Cloud Library) \cite{itseez2015opencv,Rusu_ICRA2011_PCL}.
It's popularity is due to it's relatively low computational cost and
the fact that it is Euclidean invariant.

\subsection{\label{subsec:Explicit_Plane_Fitting}Explicit Plane Fitting to Point
Cloud Data}

The explicit formulation seeks to minimize the square of the distance
between the measured data points and the estimated planar model with
respect to the plane at $Z=0$ or the $XY-$plane as shown by the
objective function (\ref{eq:explicit_fit_std}).

\begin{equation}
\epsilon(a,b,c)=\min_{a,b,c}\sum_{i=1}^{N}\left(aX_{i}+bY_{i}+c-Z_{i}\right)^{2}\label{eq:explicit_fit_std}
\end{equation}

Note that, in contrast to equation (\ref{eq:implicit_plane_fitting_error}),
this planar model has explicit form $f(X,Y)=Z$. Minimization of this
error seeks to estimate the Z-offset, $c$, and slope of the plane
with respect to the $x$-axis, $a$, and $y$-axis, \textbf{$b$.}

To optimize this function, we re-write the objective function as a
quadratic matrix-vector product by defining the vector $\mathbf{{\alpha}}=[\begin{array}{ccc}
a & b & c\end{array}]^{t}$ as the vector of planar coefficients, the vector $\mathbf{b}=[\begin{array}{cccc}
Z_{0} & Z_{1} & \ldots & Z_{N}]^{t}\end{array}$ which denotes the target depth values and the matrix $\mathbf{{M}}$
as the matrix of planar monomials formed from the 3D $(X,Y,Z)$ surface
data having $i^{th}$ row $\mathbf{{M}}_{i}=[\begin{array}{ccc}
X_{i} & Y_{i} & 1\end{array}]$. Using this notation, the optimization function becomes:
\[
\epsilon(\mathbf{{\alpha}})=\min_{\mathbf{\alpha}}\left(\mathbf{{\alpha}}^{t}\mathbf{M}^{t}\mathbf{M}\mathbf{{\alpha}}-2\mathbf{{\alpha}}^{t}\mathbf{M}^{t}\mathbf{b}+\mathbf{b}^{t}\mathbf{b}\right)
\]

Taking the derivative of the error function and setting it to zero
provides:

\begin{equation}
\frac{d\epsilon(\mathbf{{\alpha}})}{d\mathbf{{\alpha}}}=\mathbf{M}^{t}\mathbf{M}\mathbf{{\alpha}}-\mathbf{M}^{t}\mathbf{b}=0\label{eq:derivative of error-1}
\end{equation}

A solution to this system of equations is obtained via $\mathbf{\widehat{{\alpha}}}=\left(\mathbf{M}^{t}\mathbf{M}\right)^{-1}\mathbf{M}^{t}\mathbf{b}$.
Again, $\mathbf{M}^{t}\mathbf{M}$ is a symmetric matrix and, for
the monomials $\mathbf{{M}}_{i}=[\begin{array}{ccc}
X_{i} & Y_{i} & 1\end{array}]$, the elements of the matrix-vector product are

\begin{equation}
\mathbf{M}^{t}\mathbf{M}=\sum_{i=1}^{N}\left[\begin{array}{ccc}
X_{i}^{2} & X_{i}Y_{i} & X_{i}\\
X_{i}Y_{i} & Y_{i}^{2} & Y_{i}\\
X_{i} & Y_{i} & 1
\end{array}\right],\mathbf{M}^{t}\mathbf{b}=\sum_{i=1}^{N}Z_{i}\left[\begin{array}{c}
X_{i}\\
Y_{i}\\
1
\end{array}\right]\label{eq:XYZ_scattermatrix-1}
\end{equation}

Researchers have found that explicit fitting methods perform similarly
to implicit methods when measurements of the surface are normally
distributed. However, there is bias associated with the explicit fitting
objective function. Specifically, errors for explicit fits are are
measured along the $Z$ axis. This has the effect that, in contrast
to the implicit fitting approach, the estimated coefficients are not
Euclidean invariant. For this reason explicit fitting methods are
less popular for point cloud data.

\section{Methodology}

Our new approach for 3D plane fitting suggests a rearrangement of
the standard plane fitting error functions for RGBD sensor data. While
standard planar representations adopt an implicit or explicit form
of equation $aX+bY+cZ+d=0,$ our approach substitutes the RGBD 3D
reconstruction equations of \S~\ref{eq:RGBD_reconstruction} for
the variables $X$ and $Y$ and then simplifies the equation to forms
that require less computational cost to solve.

\begin{figure}
\begin{centering}
\includegraphics[scale=0.33]{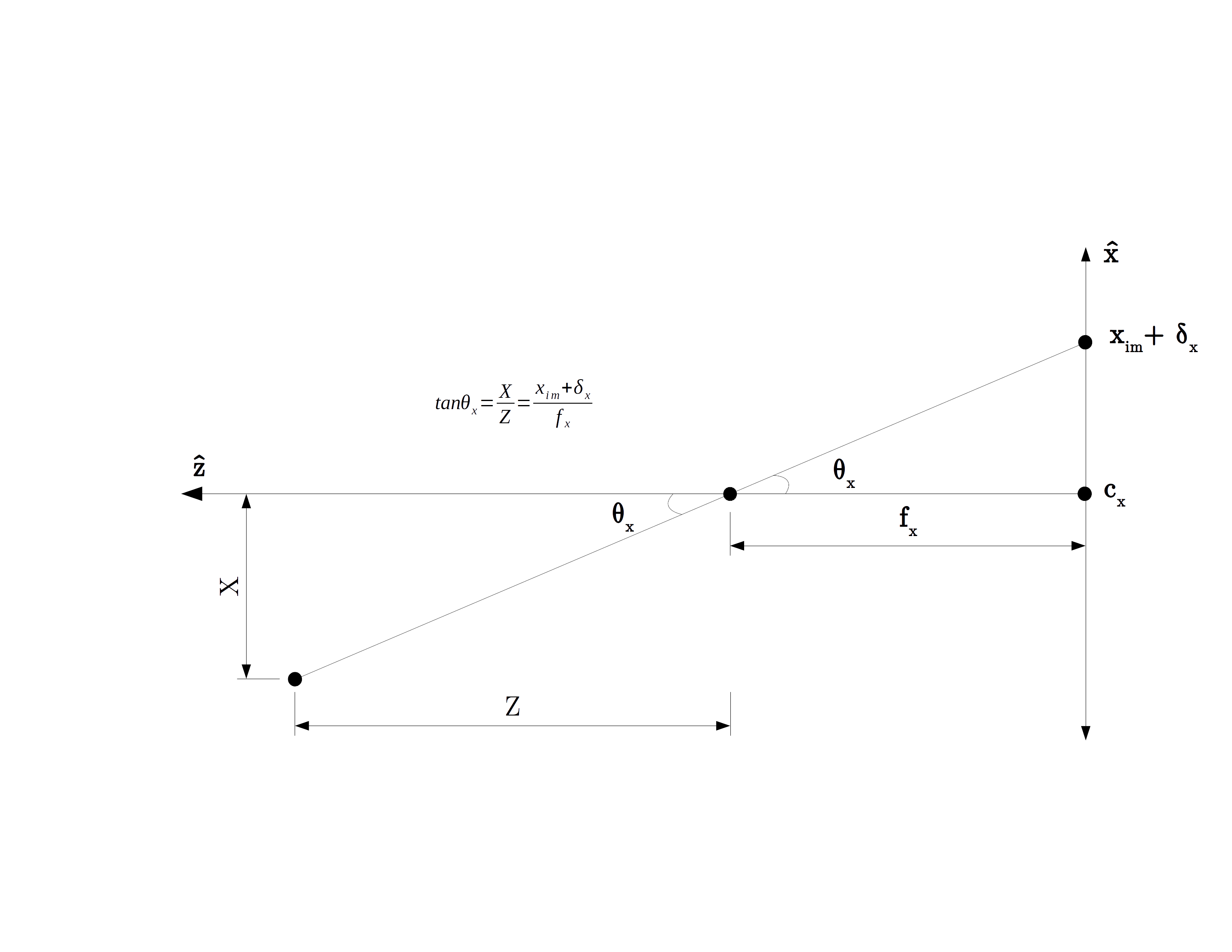}
\par\end{centering}
\caption{\label{fig:tan_theta} Note the relationship formed by the tangent
of the angle \ensuremath{\theta}\protect\textsubscript{x}. A similar
relationship exists in the y direction.}
\end{figure}

We start by investigating the RGBD reconstruction equations and grouping
the parameters into two sets: (1) the RGBD camera calibration parameters
$\left\{ f_{x},f_{y},c_{x}c_{y},\delta_{x},\delta_{y}\right\} $ and
(2) the depth measurement $Z$. Fig. \ref{fig:tan_theta} shows the
geometry of the RGBD depth measurement using these parameters. Here,
we show the angle, $\theta_{x}$, that denotes the angle between the
optical axis and line passing through the RGBD image pixel $(x,y)$
when viewed from the top-down. A similar angle, $\theta_{y}$, is
obtained using the same geometric setup from a side-view. Hence, we
make the substitutions shown in (\ref{eq:RGBD_reconstruction}) for
the terms of the reconstruction equations resulting in the new reconstruction
equations (\ref{eq:RGBD_reconstruction-1-1}):

\begin{equation}
\begin{array}{ccc}
\tan\theta_{x} & = & (x+\delta_{x}-c_{x})/f_{x}\\
\tan\theta_{y} & = & (y+\delta_{y}-c_{y})/f_{y}\\
Z & = & Z
\end{array}\label{eq:RGBD_reconstruction-1}
\end{equation}

\begin{equation}
\begin{array}{ccc}
X & = & Z\tan\theta_{x}\\
Y & = & Z\tan\theta_{y}\\
Z & = & Z
\end{array}\label{eq:RGBD_reconstruction-1-1}
\end{equation}

Back-substitution into the plane equation gives $aZ\tan\theta_{x}+bZ\tan\theta_{y}+cZ+d=0$.
We now multiply this equation by $1/Z$ to arrive at the equation
(\ref{eq:RGBD_space_plane_eq}):
\begin{equation}
a\tan\theta_{x}+b\tan\theta_{y}+c+\frac{d}{Z}=0\label{eq:RGBD_space_plane_eq}
\end{equation}
This re-arrangement of terms is the key to the contributions proposed
in this article. The benefits of this rearrangement are as follows:
\begin{itemize}
\item Coefficients $a,b,c$ are functions of \emph{only the RGBD camera
calibration parameters} which determine, for each RGBD image pixel,
the values of $\tan\theta_{x}$ and $\tan\theta_{y}$.
\item Coefficient $d$ is \emph{only }a linear function of the measured
depth (inverse depth).
\item Objective functions defined using equation (\ref{eq:RGBD_space_plane_eq})
separate uncertainty in camera calibration parameters (coefficients
$a,b,c$) from uncertainty in the measured data $d$.
\end{itemize}
In contrast to prior work \cite{6233120,TitusJiaJieTang2011} which
adopts the planar parameterization of equation (\ref{eq:prior_work_plane_equation}),

\begin{equation}
\frac{af\beta}{d}\frac{X}{Z}+\frac{bf\beta}{d}\frac{Y}{Z}+\frac{cf\beta}{d}=\frac{f\beta}{Z}\label{eq:prior_work_plane_equation}
\end{equation}

where $\beta$ is the horizontal baseline parameter, and $f$ is the
(common) focal length, our parameterization isolates measured depth
data from the RGBD camera calibration parameters. This provides computational
benefits heretofore not discussed in published work.

\subsection{Implicit Plane Fitting to RGBD Range Data}

For implicit plane fitting using the re-organized equation (\ref{eq:RGBD_space_plane_eq}),
the new monomial vector becomes $\mathbf{{M}}_{i}=[\begin{array}{cccc}
\tan\theta_{x_{i}} & \tan\theta_{y_{i}} & 1 & \frac{1}{Z_{i}}\end{array}]$ and the scatter matrix has elements:

\begin{equation}
\mathbf{M}^{t}\mathbf{M}=\sum_{i=1}^{N}\left[\begin{array}{cccc}
\tan\theta_{x_{i}}{}^{2}\\
\tan\theta_{x_{i}}\tan\theta_{y_{i}} & \tan\theta_{y_{i}}^{2}\\
\tan\theta_{x_{i}} & \tan\theta_{y_{i}} & 1\\
\frac{\tan\theta_{x_{i}}}{Z_{i}} & \frac{\tan\theta_{y_{i}}}{Z_{i}} & \frac{1}{Z_{i}} & \frac{1}{Z_{i}^{2}}
\end{array}\right]\label{eq:tan_theta_scattermatrix}
\end{equation}

where the symmetric elements of the upper-triangular matrix have been
omitted to preserve space. It is important to note that only 4 elements
of this matrix, $[\begin{array}{cccc}
\frac{\tan\theta_{x_{i}}}{Z_{i}} & \frac{\tan\theta_{y_{i}}}{Z_{i}} & \frac{1}{Z_{i}} & \frac{1}{Z_{i}^{2}}\end{array}]$, depend on the measured depth data and, as such, this matrix requires
less (\textasciitilde{}50\% less) operations to compute. The key contribution
of this article is to make the observation that the upper left 3x3
matrix of $\mathbf{M}^{t}\mathbf{M}$ \emph{does not depend upon the
measured sensor data! }As such, once the calibration parameters of
the camera are known, i.e., $(f_{x},f_{y})$, $(c_{x},c_{x})$, $(\delta_{x},\delta_{y})$,
\emph{determine all of the elements of $\mathbf{M}^{t}\mathbf{M}$
and may be pre-computed before any data is measured} with the exception
of the 4 elements in the bottom row.

\subsection{Explicit Plane Fitting to RGBD Range Data}

For explicit plane fitting using the re-organized equation (\ref{eq:RGBD_space_plane_eq}),
the revised error function is shown below:

\begin{equation}
\epsilon(a,b,c,d)=\min_{a,b,c}\sum_{i=1}^{N}\left(a\tan\theta_{x_{i}}+b\tan\theta_{y_{i}}+c-\frac{1}{Z_{1}}\right)^{2}\label{eq:RGBD_explicit_fit}
\end{equation}

Let $\mathbf{{\alpha}}=[\begin{array}{ccc}
a & b & c\end{array}]^{t}$ denote the vector of explicit plane coefficients, $\mathbf{b}=1/[\begin{array}{cccc}
Z_{0} & Z_{1} & \ldots & Z_{N}]^{t}\end{array}$ denote the target depth values and $\mathbf{{M}}$ denote the matrix
of planar monomials formed from the 3D $(X,Y,Z)$ surface data having
$i^{th}$ row $\mathbf{{M}}_{i}=[\begin{array}{ccc}
\tan\theta_{x_{i}} & \tan\theta_{y_{i}} & 1\end{array}]$. 

Then equation (\ref{eq:tan_theta_scattermatrix-1}) shows the scatter
matrix, $\mathbf{M}^{t}\mathbf{M}$, needed to estimated the explicit
plane coefficients where the symmetric elements of the upper-triangular
matrix have been omitted to preserve space. 

\begin{equation}
\mathbf{M}^{t}\mathbf{M}=\sum_{i=1}^{N}\left[\begin{array}{ccc}
\tan\theta_{x_{i}}{}^{2}\\
\tan\theta_{x_{i}}\tan\theta_{y_{i}} & \tan\theta_{y_{i}}^{2}\\
\tan\theta_{x_{i}} & \tan\theta_{y_{i}} & 1
\end{array}\right]\label{eq:tan_theta_scattermatrix-1}
\end{equation}

It is important to note that \emph{none of the} \emph{elements of
the scatter matrix depend on the measured depth data and, as such,
this matrix requires a constant number of operations to compute for
each measured image, i.e., it can be pre-computed given the RGBD camera
parameters. }Hence, explicit plane fitting in the RGBD range space
requires only computation of the vector $\mathbf{b}=1/[\begin{array}{cccc}
Z_{0} & Z_{1} & \ldots & Z_{N}]\end{array}$ for each range image and the best-fit plane is given by a single
matrix multiply: $\mathbf{\widehat{{\alpha}}}=\left(\mathbf{M}^{t}\mathbf{M}\right)^{-1}\mathbf{M}^{t}\mathbf{b}$.
Where the value of $\mathbf{M}^{t}\mathbf{b}$ is given below:

\[
\mathbf{M}^{t}\mathbf{b}=\sum_{i=1}^{N}\frac{1}{Z_{i}}\left[\begin{array}{c}
\tan\theta_{x_{i}}\\
\tan\theta_{y_{i}}\\
1
\end{array}\right]
\]

\subsection{\label{subsec:Integral_Images}Computational Savings via Integral
Images}

Integral images, first introduced in \cite{lewis1995fast} and popularized
in the vision community by the work \cite{990517}, computes the cumulative
distribution of values within an image. This technique is often applied
for the purpose of efficiently computing sums over arbitrary rectangular
regions of an image. This is often the case for normal estimation
in range images. Here, integral images are computed for the $(X,Y,Z)$
values or perhaps for all values of the scatter matrix of equation
(\ref{eq:XYZ_scattermatrix}). Computation of an integral image for
a scalar value, e.g., intensity, requires 4 operations per pixel,
i.e., for an image of $N$ pixels it has computational cost of $t=4N$
operations. More generally, computation of an integral image for an
arbitrary function of the image pixel data $f(I(x,y))$ having computational
cost $C$ has computational cost to $t=N(C+4)$. 

\subsection{\label{subsec:Implicit-Fitting-Cost}Implicit Fitting Computational
Cost Analysis}

Assuming that the images $(\tan\theta_{x}(x,y),\tan\theta_{y}(x,y))$
are pre-computed, the function $f()$ needed to compute $X$ and $Y$
coordinates requires one operation (a single multiplication) for each
pixel, i.e., $C=1$ for these matrix elements. Likewise computation
of $(X^{2},Y^{2},Z^{2},XY,XZ,YZ)$ from the $X$ and $Y$ coordinates
requires an additional operation (again, multiplication), i.e., $C=1$
for these matrix elements. Since the depth is directly measured, computation
of the depth integral image has no additional computational cost.
Hence, computation of the integral images for the 9 unique elements
of the scatter matrix of equation (\ref{eq:XYZ_scattermatrix}) requires
$t_{std.}=8N(1+4)+4N=44N$ operations. Note that every element (except
the constant) in this matrix is a function of the measured depth image
data $Z$.

In contrast, the scatter matrix of equation (\ref{eq:tan_theta_scattermatrix})
contains only 4 elements that depend on the depth data: $(\frac{\tan\theta_{x_{i}}}{Z_{i}},\frac{\tan\theta_{y_{i}}}{Z_{i}},\frac{1}{Z_{i}},\frac{1}{Z_{i}^{2}})$.
As before, the function $f()$ needed to compute each term requires
a single operation. Hence, integral images for all elements of the
scatter matrix of equation (\ref{eq:tan_theta_scattermatrix}) requires
$t_{new}=4N(1+4)=20N$ operations to compute where we do not include
the cost of computing the constant integral images for the 5 elements
of the matrix that do not depend on measured depth data (a cost of
$t=20N$).

Hence, analysis indicates that the computational cost of an integral-image
based solution for the approach prescribed by equation (\ref{eq:XYZ_scattermatrix})
requires more than twice as many operations ($\frac{t_{std.}}{t_{new}}=2.2$)
than the approach prescribed by equation (\ref{eq:tan_theta_scattermatrix}).

Given that RGBD image data is captured at a resolution of $N=640\times480\approx307k$
and framerate of $30$ images/second these computational savings may
significantly affect the runtime of real-time image processing algorithms
for this class of image sensors.

In a naive implementation using a square window containing $M$ pixels,
each element of the scatter matrix requires ${\cal O}(M^{2})$ operations
and the cost of computing these elements for an entire image of pixel
data is ${\cal O}(NM^{2})$. In this case, the computational costs
are $t_{std.}=8NM^{2}$ and $t_{new}=4NM^{2}$ (or a ratio $\frac{t_{std.}}{t_{new}}=2$)
to implement equations (\ref{eq:XYZ_scattermatrix}) and (\ref{eq:tan_theta_scattermatrix})
respectively.

\subsection{\label{subsec:Explicit-Fitting-cost}Explicit Fitting Computational
Cost Analysis}

Following the previous analysis and assuming an integral image implementation,
the complexity of explicit fitting via equation (\ref{eq:XYZ_scattermatrix-1})
is $t_{std}=7N(4+1)+4N=39N$ and that of explicit fitting as in equation
(\ref{eq:tan_theta_scattermatrix-1}) is $t_{new}=15N$. This suggests
even larger computational savings than those found for implicit fitting,
e.g., ($\frac{t_{std.}}{t_{new}}=2.6$), which may prove extremely
beneficial for RGBD depth image processing, e.g., real-time surface
analysis algorithms.

Note that the computational complexity discussions for both implicit
and explicit planar fits do not include the computational cost of
computing eigenvectors or the matrix inverse respectively. In both
cases the complexity of this operation is ${\cal O}(D^{3})$ where
$D=3$ and this cost is not considered as it is common to all discussed
methods.

\section{Results}

A series of timing experiments were performed using MATLAB on a 2.7
GHz i7-2620M processor to evaluate the runtime of the proposed fitting
methods when compared to standard approaches. First, a 640x480 depth
senor was modeled in terms of the parameters $(f_{x},f_{y})$, $(c_{x},c_{x})$,
and $(\delta_{x},\delta_{y})$. A series of 3D rays was formed emanating
from the center of each pixel through the focal point of the virtual
camera. The intersection of these rays and an arbitrary 3D plane was
computed and the resulting sample points perturbed with Gaussian noise
in the direction of the measurement. Values for $\tan\theta_{x}(x,y)$
and $\tan\theta_{y}(x,y)$ were also computed during this process.
These generated sample points and values for $\tan\theta_{x}(x,y)$,
$\tan\theta_{y}(x,y)$ were then used in the following experiments. 

\subsection{Integral Image Implementation}

Often, multiple planes are fit to subsets of a single frame of depth
image data. It can be advantageous then, to employ integral images
to more efficiently compute sums over arbitrary rectangular image
regions as described in \S~\ref{subsec:Integral_Images}. Using
this integral image implementation, fitting in RGBD range space is
superior to the standard approach in terms of runtime. 

\begin{figure}
\begin{centering}
\includegraphics[width=7.75cm]{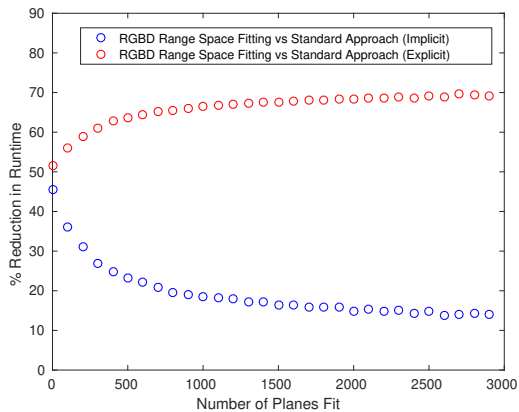}
\par\end{centering}
\caption{\label{fig:Integral_Percent_Faster_vs_Iterations}Comparison of integral
image RGBD range space plane fitting approaches vs integral image
standard approaches for a single frame of depth image data. Computation
of the integral images used in each method is included, with RGBD
range space fitting approaches taking half as much time as time to
compute the required integral images as their standard counterparts.}
\end{figure}

Fitting in RGBD range space has an advantage using this implementation
due to the fact that integral images involving $\tan\theta_{x}(x,y)$
and $\tan\theta_{y}(x,y)$ can be pre-computed and do not change for
a given sequence of range images. Therefore, fewer integral images
need to be computed on a per image basis when compared to the standard
approach using a similar integral image implementation. Fig. \ref{fig:Integral_Percent_Faster_vs_Iterations}
compares the integral image implementations of both implicit and explicit
RGBD range space fitting methods to their standard counterparts. For
all integral methods, the time taken to compute the required integral
images and fit a varying number of planes was compared. In the implicit
case, integral image computation time was reduced by 45\% and individual
plane fitting time was reduced by \textasciitilde{}15\% compared to
the standard approach. Using explicit fitting, integral image computation
time was reduced by 52\% and individual plane fitting time was reduced
by \textasciitilde{}70\%. The reduction in integral image computation
time is consistent with the analysis in \S~\ref{subsec:Implicit-Fitting-Cost}
and \S~\ref{subsec:Explicit-Fitting-cost}.
\begin{figure}
\begin{centering}
\includegraphics[width=7.75cm]{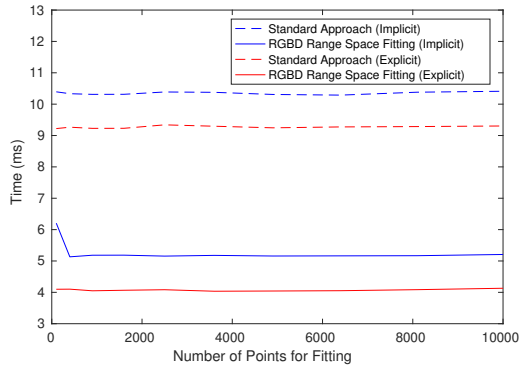}
\par\end{centering}
\caption{\label{fig:Integral_Times_vs_Points}Runtime comparison of implicit
and explicit fitting approaches using the integral image implementation.
This data includes time taken to compute these integral images and
perform a single plane fit with a varying number of points. Note that
the time taken to perform these operations is constant despite the
increasing number of points used in each fit.}
\end{figure}
\begin{figure*}
\begin{centering}
\includegraphics[width=5.25cm]{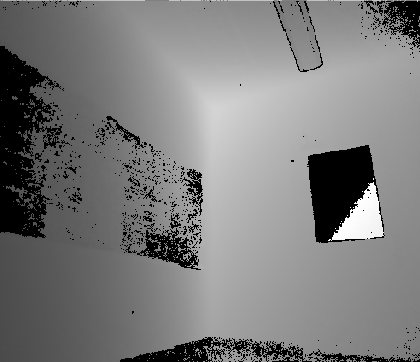} \includegraphics[width=5.25cm]{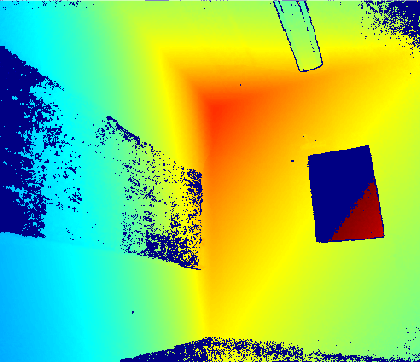}
\includegraphics[width=5.25cm]{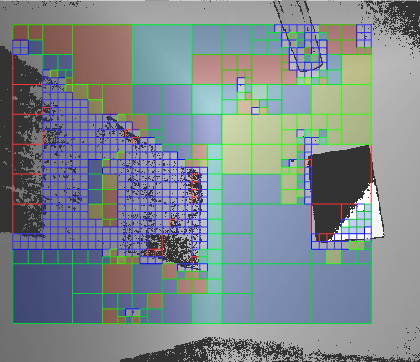}
\par\end{centering}
\caption{\label{fig:Planar_Segmentation}Depth image planar segmentation of
a room corner featuring a blackboard and window. Depth image (left),
color mapped depth image (middle), and segmented image (right). The
algorithm successfully fit 268 planes on the 512x424 image. Tiles
with successful plane fits are shaded and outlined in green. Tiles
with too many invalid points or high fit error are outlined in red
and blue respectively.}
\end{figure*}

When performing implicit plane fitting in RGBD range space, precomputed
values are used to fill the scatter matrix of equation (\ref{eq:tan_theta_scattermatrix})
directly, leaving only the 4 elements that depend on $\frac{1}{Z}$
to be computed for each image. In contrast, the standard plane fitting
approach requires computation of all elements of the scatter matrix
from equation (\ref{eq:XYZ_scattermatrix}).

A similar implementation can be used for explicit plane fitting methods.
When fitting in RGBD range space, there is a distinct advantage in
terms of computation as the matrix elements of equation (\ref{eq:tan_theta_scattermatrix-1})
do not depend on the measured depth data. This matrix depends only
on the camera's intrinsic parameters, and can be pre-computed. As
shown in Fig. \ref{fig:Integral_Times_vs_Points} and \ref{fig:Integral_Percent_Faster_vs_Iterations},
these computational savings result in a 70\% speedup in fitting planes
for each measured image.

\subsection{Naive or ``Standard'' Implementation}

\begin{figure}
\begin{centering}
\includegraphics[width=7.75cm]{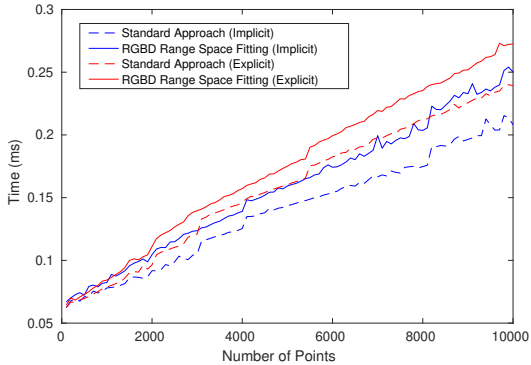}
\par\end{centering}
\caption{\label{fig:Naive_Times_vs_Points}Runtime comparison of implicit and
explicit fitting approaches using the naive implementation (without
integral image computation). For plane fits involving less than 10,000
points, explicit RGBD range space fitting is \textasciitilde{}10\%
slower and implicit RGBD range space fitting is \textasciitilde{}15\%
slower than the corresponding standard approaches.}
\end{figure}

The naive implicit fitting implementation computes the scatter matrix
elements of equations (\ref{eq:XYZ_scattermatrix}) and (\ref{eq:tan_theta_scattermatrix})
through the multiplication $\mathbf{M}^{t}\mathbf{M}$. When constructing
each monomial vector $\mathbf{{M}}$, the standard approach requires
two multiplication operations to obtain the X and Y values of the
3D points, while fitting in RGBD range space requires a single division
to obtain the values $\frac{1}{Z}$. Fit times for both methods as
a function of the number of points used is shown in Fig. \ref{fig:Naive_Times_vs_Points}.
Results show that implicit fitting in RGBD range space is slower by
\textasciitilde{}15\% compared to the standard approach when using
this naive implementation. This unexpected time difference is attributed
to the significant difference in computational cost between a multiply
operation (\textasciitilde{}1-3 clock cycles on contemporary CPUs)
and the divide operation (\textasciitilde{}10-20 clock cycles). The
large discrepancy in the cost of the division dominates and the savings
afforded by having half as many operations as discussed in \S\ref{subsec:Implicit-Fitting-Cost}
and \S\ref{subsec:Explicit-Fitting-cost} is superseded by this additional
cost. Direct fitting using the disparity images which is directly
proportional to inverse depth, $\frac{1}{Z}$, is very likely to alleviate
these costs but investigation of these benefits is beyond the scope
of this article.

Similarly, the naive implementation of the explicit fitting approaches
described in \S~\ref{subsec:Explicit_Plane_Fitting} compute the
matrices of equations (\ref{eq:XYZ_scattermatrix-1}) and (\ref{eq:tan_theta_scattermatrix-1})
through the multiplication $\mathbf{M}^{t}\mathbf{M}$. Using the
standard explicit fitting approach, the X and Y values of the 3D points
must be computed to construct the monomial vector. Explicit fitting
in RGBD range space requires no additional computation to construct
the monomial vector, as values for $\tan\theta_{x}(x,y)$ and $\tan\theta_{y}(x,y)$
can be pre-computed. However, the vector $\mathbf{b}=1/[\begin{array}{cccc}
Z_{0} & Z_{1} & \ldots & Z_{N}]\end{array}$ in this case, so values for $\frac{1}{Z}$ must be computed. We observed
explicit fitting in RGBD range space to be \textasciitilde{}10\% slower
than the standard explicit approach using this naive implementation.

\subsection{Integral Image vs. Standard Implementation Analysis}

All implicit and explicit plane fitting methods were compared by examining
runtime vs number of planes fit for a single depth image. Each plane
was fit repeatedly to a 50x50 pixel subset of the overall image totaling
2500 points. For approaches involving integral images, computation
of integral images was included in the overall runtime. 

\begin{figure}[h]
\begin{raggedright}
\subfloat[]{\begin{centering}
\includegraphics[width=8cm]{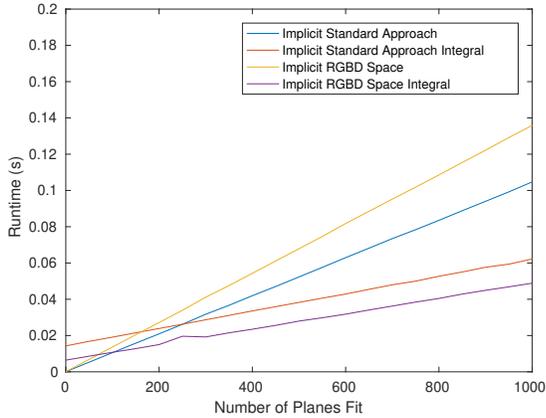}
\par\end{centering}

}
\par\end{raggedright}
\begin{raggedright}
\subfloat[]{\begin{centering}
\includegraphics[width=8cm]{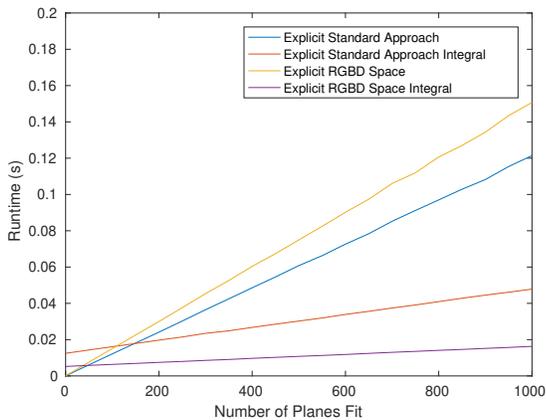}
\par\end{centering}
}
\par\end{raggedright}
\caption{\label{fig:Runtime_Comparison} (a,b) compare performance of implicit
(a) and explicit (b) plane fitting methods for a single depth image.
As the number of planes fit per image grows larger, the integral image
implementation of RGBD range space fitting quickly surpasses all other
plane fitting methods in terms of speed.}
\end{figure}

As shown in Fig. \ref{fig:Runtime_Comparison} in both the implicit
and explicit cases, fitting in RGBD range space using an integral
image implementation is faster than all other methods. Fitting 200
planes, implicit integral image RGBD space fitting takes 37\% less
time than the standard integral image approach and 28\% less time
than naive standard fitting. Explicit integral image RGBD range space
fitting takes 62\% less time than the standard explicit integral image
approach and 50\% less time than even implicit integral image RGBD
space fitting. 

\subsection*{Planar Segmentation Application}

The integral image implementations of all plane fitting methods were
compared by their application in a planar segmentation algorithm.
The algorithm takes in a depth image and divides into an equally sized
square grid. For each grid element, a plane is fit using each type
of integral image plane fitting approach- integral image implementations
of the standard and RGBD space fitting methods. 

If the error in the fit is above a threshold value, the grid element
is subdivided into 4 equally smaller elements, each with half the
width and height of the parent. This subdivision and plane fitting
process is then repeated for these child grid elements, up to a maximum
number of subdivisions. K-means clustering was then applied to planar
coefficients to classify and label similar planes. This process results
in the planar segmentation of the depth image as shown in Fig. \ref{fig:Planar_Segmentation}. 

For this algorithm, time spent plane fitting and computing integral
images for 30 frames of depth image data was recorded. As previous
analysis indicated, the integral image implementation of implicit
RGBD range space fitting demonstrated significant time savings compared
to its standard counterpart, taking 22\% less time. In the explicit
case, integral image RGBD range space fitting produced a 32\% reduction
in runtime compared to standard integral image fitting. Between the
two fitting formulations, the explicit RGBD range space fitting approach
was the fastest, taking 66\% less time than implicit RGBD range space
fitting. Extrapolation of these results indicate that over a five
minute quadcopter flight the proposed algorithm would save over 30
seconds of processing time which may then be used for other on-board
computational tasks.

\section{Conclusion}

The problem of real time surface fitting has shown itself to be an
important challenge in the area of computer vision. In this work,
we have introduced computationally efficient methods for least squares
3D plane fitting, which outperform the standard approaches when implemented
using integral images. These methods improve upon existing approaches
by separating the traditional fitting problem into components including
the intrinsic camera parameters, which can be considered constant,
and quantities related to the disparity of the incoming depth data.

This article proposed a reformulation of standard 3D plane fitting
which, when coupled with integral imaging techniques, significantly
reduces the computational cost associated with plane fitting to 3D
point cloud data. We have also demonstrated the practical advantages
in employing such methods in terms of a planar segmentation algorithm,
which showed a substantial reduction in runtime consistent with other
experimental results. We feel these results show promise for application
of the proposed surface fitting methods in real time RGBD vision systems. 

Future work may investigate a surface fitting approach dealing with
the disparity data directly. After all, the quantity $\frac{1}{Z}$
is directly proportional to the sensed disparity. If depth information
is not explicitly required, additional computational savings can be
realized by working directly from measured disparities which may also
eliminate the cost of depth computation which typically is performed
by the sensor driver. It is also likely that surfaces of higher order
can be estimated using a similar surface fitting formulation as proposed
in this work, although it remains to be seen if such approaches would
provide increased performance.

\bibliographystyle{ieeetr}
\bibliography{2017_SeCon_RGBD_RealTimeSurfaceFitting}

\begin{thebibliography}{10}

\bibitem{6547194}
J.~Han, L.~Shao, D.~Xu, and J.~Shotton, ``Enhanced computer vision with
  microsoft kinect sensor: A review,'' {\em IEEE Transactions on Cybernetics},
  vol.~43, pp.~1318--1334, Oct 2013.

\bibitem{Holz11real-timeplane}
D.~Holz, S.~Holzer, R.~B. Rusu, and S.~Behnke, ``Real-time plane segmentation
  using rgb-d cameras,'' in {\em In Proc. of the 15th RoboCup Int. Symp}, 2011.

\bibitem{s120201437}
K.~Khoshelham and S.~O. Elberink, ``Accuracy and resolution of kinect depth
  data for indoor mapping applications,'' {\em Sensors}, vol.~12, no.~2,
  p.~1437, 2012.

\bibitem{Fitzgibbon:1999:DLS:302943.302950}
A.~Fitzgibbon, M.~Pilu, and R.~B. Fisher, ``Direct least square fitting of
  ellipses,'' {\em IEEE Trans. Pattern Anal. Mach. Intell.}, vol.~21,
  pp.~476--480, May 1999.

\bibitem{Taubin:Fitting:1991}
G.~Taubin, ``Estimation of planar curves, surfaces and nonplanar space curves
  defined by implicit equations, with applications to edge and range image
  segmentation,'' {\em IEEE Transactions on Pattern Analysis and Machine
  Intelligence}, vol.~13, no.~11, pp.~1115--1138, 1991.

\bibitem{Trucco:1998:ITC:551277}
E.~Trucco and A.~Verri, {\em Introductory Techniques for 3-D Computer Vision}.
\newblock Upper Saddle River, NJ, USA: Prentice Hall PTR, 1998.

\bibitem{itseez2015opencv}
Itseez, ``Open source computer vision library.''
  \url{https://github.com/itseez/opencv}, 2015.

\bibitem{Rusu_ICRA2011_PCL}
R.~B. Rusu and S.~Cousins, ``{3D is here: Point Cloud Library (PCL)},'' in {\em
  {IEEE International Conference on Robotics and Automation (ICRA)}},
  (Shanghai, China), May 9-13 2011.

\bibitem{6233120}
C.~Erdogan, M.~Paluri, and F.~Dellaert, ``Planar segmentation of rgbd images
  using fast linear fitting and markov chain monte carlo,'' in {\em 2012 Ninth
  Conference on Computer and Robot Vision}, pp.~32--39, May 2012.

\bibitem{TitusJiaJieTang2011}
T.~J.~J. Tang, W.~L.~D. Lui, and W.~H. Li, ``A lightweight approach to 6-dof
  plane-based egomotion estimation using inverse depth,'' in {\em Australasian
  Conference on Robotics and Automation}, 2011.

\bibitem{lewis1995fast}
J.~P. Lewis, ``Fast template matching,'' in {\em Vision interface}, vol.~95,
  pp.~15--19, 1995.

\bibitem{990517}
P.~Viola and M.~Jones, ``Rapid object detection using a boosted cascade of
  simple features,'' in {\em Proceedings of the 2001 IEEE Computer Society
  Conference on Computer Vision and Pattern Recognition. CVPR 2001}, vol.~1,
  pp.~I--511--I--518 vol.1, 2001.

\end{thebibliography}

\end{document}